# Data-Driven Model Discrimination of Switched Nonlinear Systems with Temporal Logic Inference

Zeyuan Jin*, Nasim Baharisangari*, Zhe Xu, and Sze Zheng Yong

*Abstract*— This paper addresses the problem of data-driven model discrimination for unknown switched systems with unknown linear temporal logic (LTL) specifications, representing tasks, that govern their mode sequences, where only sampled data of the unknown dynamics and tasks are available. To tackle this problem, we propose data-driven methods to over-approximate the unknown dynamics and to infer the unknown specifications such that both set-membership models of the unknown dynamics and LTL formulas are guaranteed to include the ground truth model and specification/task. Moreover, we present an optimization-based algorithm for analyzing the distinguishability of a set of learned/inferred model-task pairs as well as a model discrimination algorithm for ruling out model-task pairs from this set that are inconsistent with new observations at run time. Further, we present an approach for reducing the size of inferred specifications to increase the computational efficiency of the model discrimination algorithms.

## I. INTRODUCTION

In cyber-physical systems (CPS) and multi-agent systems (MAS), various sensors, actuators, and subsystems/agents are interconnected, where each component in CPS or each agent in MAS may perform a distinct task while satisfying a distinct dynamical model. Distinguishing the components or agents and the tasks from each other becomes crucial for various purposes. For instance, to detect faults and minimize the risk of serious damages or failures in CPS, it is necessary to identify which component has abnormal functionality, while in MAS, it is desirable to detect the tasks/intents of other dynamic agents. For the purpose of distinguishing amongst different model behaviors, we can leverage the differences that are the result of different system dynamics or specifications, where the *system dynamics* governs the evolution of system states based on physical laws and the *specification* governs the temporal evolution of their system modes corresponding to desired tasks or rules. The task specifications are often expressed using temporal logic formulas, such as linear temporal logic (LTL) formulas, which is a highly expressive language capable of providing formal yet easily understandable descriptions of system behaviors [1], [2]. Temporal logic formulas are widely used across many fields, including control synthesis and complex robotic applications, anomaly detection in underlying systems, and specification, recognition, and interpretation in dynamic environments [3]–[11].

Thus, discriminating among different models can be expedited by considering both the specifications and system dynamics in contrast to only utilizing the specifications or system dynamics alone. However, the exact specifications and system dynamics are usually not available, which makes this problem more challenging and interesting.

**Related Work:** Model discrimination is the task of distinguishing between models based on a finite sequence of measured input-output data [12], [13], [14]. This problem can be approached by using a modeling invalidation framework, which seeks to determine whether the observed input-output data is consistent with any member of the set of valid models [15]. The problem of model invalidation has been studied for different types of systems, including linear parameter varying systems [16], [17], nonlinear systems [18], uncertain systems [19], switched auto-regressive models [20], and switched affine systems [13], [21], [22]. Additionally, to examine the detectability of the models, $T$-distinguishability (or $T$-detectability) is introduced in [13], [22] to find upper bounds on the required time horizon $T$ to distinguish one model from the other, if such a $T$ exists. The notion of $T$-distinguishability is closely related to the concept of state/mode distinguishability of switched linear systems [23], [24], finite-state systems [25] and switched nonlinear systems [26]. Recently, model discrimination and fault detection using temporal logics specifications have also gained attention. Jiang *et. al.* [27] conducted failure diagnosis by specifying the fault as an LTL formula; and Yang *et. al.* [28] combined switched affine systems and LTL formulas to constrain the switching modes so that the detectability horizon $T$ can be reduced. Niu *et. al.* [29] further extended the work in [28] to nonlinear switched systems, and they defined the specifications in the form of *metric/signal temporal logic* formulas. However, these approaches all rely on precise mathematical models of both the system dynamics and the temporal logic specifications.

When the system dynamics are unknown or only partially known, various learning approaches, e.g., Gaussian process regression, clustering-based methods and neural networks [30] have been developed. Further, to prevent the true model from being wrongly discriminated, set-membership learning approaches have also been proposed to over-approximate the unknown system dynamics from observed/sampled input-output data by finding *a set of known systems* that is guaranteed to contain the true model and retain most properties of interest with the unknown system dynamics [31], [32].

Z. Jin, N. Baharisangari and Z. Xu are with the School for Engineering of Matter, Transport and Energy, Arizona State University, Tempe, AZ 85287 ({zjin43,nbaharis,xzhe1}@asu.edu) and S.Z. Yong is with the Department of Mechanical and Industrial Engineering, Northeastern University, Boston, MA 02115 (s.yong@northeastern.edu). This work is supported by NSF grants CNS-1932066, CNS-1943545, and CNS 2304863.

*These authors equally contributed to this paper.

Within this context, the work in [33] proposes a recursive algorithm to compute upper and lower bounding functions for univariate Lipschitz continuous unknown dynamics while the study in [34], [35] considered the extension to multivariate functions. This technique was extended to unknown differentiable functions with bounded Jacobians in [36] and to Hölder continuous unknown dynamics in [37], while in [38], the authors further considered componentwise Hölder continuous functions where the contribution of each input to each output of the function is independently counted.

On the other hand, when the exact temporal logic formulas for the task specifications are unknown, the formulas can be inferred from traces of underlying systems and utilized for different purposes [6], [39]–[53]. For example, LTL task specifications were inferred/learned from demonstrations of a task using Bayesian inference approach to address the problem of acceptability of task execution, as presented in [54]. Moreover, in [55], both formal grammar and temporal logic have been used for mining the structure and the parameters of a *signal temporal logic* specification from a set of unlabeled trajectories while in [56]–[58], formal grammar and temporal logic have been used for automated recognition of complex human activities and the task of run time verification. Nonetheless, the works mentioned above do not consider the learning and inference of the unknown system dynamics and task specifications simultaneously, which we find to be synergistic for the purpose of model discrimination.

**Contributions:** In this paper, we aim to solve the problem of data-driven model discrimination among a set of unknown models with unknown switched system dynamics and unknown linear temporal logic (LTL) specifications that govern their mode sequences (representing tasks), where we only have access to data of the unknown dynamics and specifications. To address this challenge, we propose a data-driven set-membership method such that the learned dynamics model over-approximates the unknown dynamics and the ground truth LTL specification/task is included in the set of inferred specifications. Furthermore, we analyze the distinguishability of the learned data-driven model-task pairs by introducing an optimization-based algorithm that can find an upper bound on the required time horizon $T$ to guarantee the discrimination/separation of the model-task pairs, as well as propose a model discrimination algorithm that can rule out learned model-task pairs that are inconsistent with new data/observations at run time. Moreover, we propose a complexity reduction framework for reducing the size of the inferred LTL formula by leveraging prior information, when available, that can increase the computation efficiency in model discrimination algorithms. Finally, we demonstrate the effectiveness of our proposed methods via illustrative examples of intent estimation for vehicles and single link robot arms. Our results indicate that the synergistic learning of both LTL specifications and system dynamics can significantly accelerate data-driven model discrimination.

**Paper Organization:** The paper is organized as follows. Sections II and III describe the necessary definitions and models considered in this paper and formulate the problems we aim to address. Section IV introduces the proposed method to find an over-approximation of the system dynamics. In Section V, we propose an algorithm for inferring a set of compatible LTL specifications from data, and then, we propose a framework and algorithm for reducing the size of the inferred specification. In Section VI, we propose algorithms for analyzing model detectability/distinguishability and for discriminating among models. Simulation results are presented in Section VII to demonstrate the effectiveness of our approach and Section VIII concludes the paper.

## II. PRELIMINARIES

In this section, we introduce the mathematical concepts and notations used throughout this paper.

### A. Notations

$\|v\|_p$ for $p = \{1, \infty\}$ denotes the $p$-norm of a vector $v \in \mathbb{R}^n$. The set of integers from $a$ through $b$ is denoted by $\mathbb{Z}_a^b$, $\mathbb{N} \triangleq \{1, 2, ...\}$ and $|S|$ is the cardinality of the finite set $S$.

### B. Linear Temporal Logic (LTL)

First, we introduce an *atomic proposition* as a statement on a system variables denoted by $\pi$. This statement is either *True* or *False*. Let $\Sigma$ be a finite set of atomic propositions. The syntax of LTL formulas over $\Sigma$ is defined as follows.

$$\phi := \top \mid \pi \mid \neg \phi \mid \phi_1 \wedge \phi_2 \mid \mathbf{X}\,\phi \mid \phi_1\,\mathbf{U}\,\phi_2,$$

where $\pi \in \Sigma$, $\neg$ and $\wedge$ are Boolean connectives, $\mathbf{U}$ is the temporal operator "until", and $\mathbf{X}$ is the temporal operator "next". We add syntactic sugar and define temporal operators $\mathbf{G}$ ("always") and $\mathbf{F}$ ("finally"). In addition, we derive Boolean connectives $\vee$ ("disjunction") and $\rightarrow$ ("implication") from the introduced Boolean connectives. We denote the set containing all the temporal operators and all the Boolean connectives by $\mathcal{O}$.

Let $\boldsymbol{\sigma}$ be an $\omega$-word over $\Sigma$ whose $t$-th element is denoted by $\sigma_t$. We define the Boolean semantics of LTL formulas over such sequences as follows.

$$\begin{aligned}
(\boldsymbol{\sigma}, t) &\models \pi \text{ iff } \boldsymbol{\sigma}_t = \pi, \\
(\boldsymbol{\sigma}, t) &\models \neg \phi \text{ iff } (\boldsymbol{\sigma}, t) \not\models \phi, \\
(\boldsymbol{\sigma}, t) &\models \phi_1 \wedge \phi_2 \text{ iff } (\boldsymbol{\sigma}, t) \models \phi_1 \text{ and } (\boldsymbol{\sigma}, t) \models \phi_2, \\
(\boldsymbol{\sigma}, t) &\models \mathbf{X}\,\phi \text{ iff } (\boldsymbol{\sigma}, t+1) \models \phi, \\
(\boldsymbol{\sigma}, t) &\models \phi_1\,\mathbf{U}\,\phi_2 \text{ iff } \exists t' \geq t \,:\, (\boldsymbol{\sigma}, t') \models \phi_2 \\
&\qquad \text{and } \forall t'' \in [t, t') \,:\, (\boldsymbol{\sigma}, t'') \models \phi_1.
\end{aligned}$$

**Syntax DAG:** Any LTL formula can be represented as a syntax directed acyclic graph, i.e., syntax DAG. In a syntax DAG, the nodes are labeled with atomic propositions or temporal operators or Boolean connective that form an LTL formula [59]. For instance, Figure 1a shows the unique syntax DAG of the formula $(\pi_1\,\mathbf{U}\,\pi_2) \wedge \mathbf{G}(\pi_1 \vee \pi_2)$, in which the subformula $\pi_2$ is shared. Figure 1b shows the arrangement of the identifiers of each node in the syntax DAG ($k \in \{1,..,7\}$). We denote the node $k$ with its associated label by $n_k \in \mathcal{E}$ where $\mathcal{E}$ denotes the set of nodes

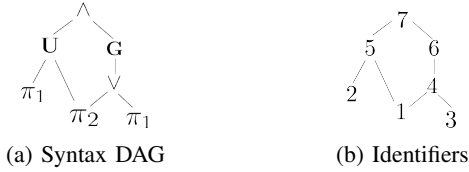

Fig. 1: Syntax DAG and identifier of syntax DAG of the formula $(\pi_1 \, \mathbf{U} \, \pi_2) \wedge \mathbf{G}(\pi_1 \vee \pi_2)$.

with their labels in the DAG that is associated with a given LTL formula.

**Size of an LTL formula:** If we represent an LTL formula by a syntax DAG, then each node corresponds to a subformula; thus, the size of an LTL formula is the number of the DAG nodes. We denote the size of an LTL formula $\phi$ by $|\phi|$ [60].

### C. Formal Grammar

**String:** If we denote an alphabet by $\Pi$, then a *string* over this alphabet is a finite sequence of symbols of $\Pi$.

**Kleene Closure:** Kleene closure, denoted by $^*$, is a unary operation on sets of symbols. If we apply the Kleene operation on the alphabet $\Pi$ (a finite set of symbols), $\Pi^*$ is the set of all strings over $\Pi$ including the empty string $\epsilon$.

**Formal Grammar:** A *formal grammar* is a set of productions (rules) that describe how to generate a string from the alphabet of a language. From a mathematical point of view, grammar is a quadruple $\langle \Pi, \mathcal{N}, \mathcal{T}, \mathcal{RU} \rangle$. In this quadruple, $\Pi$ is a finite non-empty set called *terminal alphabet*. The elements of this set are called *terminals* and are the symbols appearing in the output strings of a grammar. They are called terminals due to the fact that there is no rule in the grammar $\mathcal{RU}$ that changes them. $\mathcal{N}$ is a finite nonempty set that is disjoint from $\Pi$. The elements of this set are called *non-terminals*. Non-terminals can be replaced by terminals. $\mathcal{T} \in \mathcal{N}$ is a distinguished non-terminal which is called the "start" symbol. $\mathcal{RU}$ is a finite set of productions (rules) in the following form: $\lambda \implies \mu$, where $\lambda \in (\Pi \cup \mathcal{N})^* \mathcal{N}(\Pi \cup \mathcal{N})^*$, referred to as "head", is a string of terminals and non-terminals, and contains at least one non-terminal symbol; $\mu \in (\Pi \cup \mathcal{N})^*$, referred to as "body", is a string of terminals and non-terminals. The symbol $\implies$ is commonly used in representing a production in formal grammar.

**Context-Free Grammar (CFG):** *Context-free* grammar (CFG) is a type of formal grammar in which the head consists of a single non-terminal symbol ($\lambda \in \mathcal{N}$). CFGs are called context-free because any of the productions in $\mathcal{RU}$ can be applied to any of the non-terminals in $\mathcal{N}$ regardless of whether the non-terminals are surrounded by other symbols or not. It is common to represent the productions that are applied to a same head in the same line [61]–[63]. The symbol $\implies$ used in defining rules in formal grammar is different from the Boolean connective $\rightarrow$ ("implies") in LTL. The symbol $\implies$ is used for defining a production between the head and the body for generating a string. In addition, the symbol $|$ is commonly used for separating the "bodies" ($\mu$) following a same "head" ($\lambda$).

**Example 1.** *We define* $\Pi = \{A, a, B, b, c, C\}$ *and* $\mathcal{N} = \{\lambda_1, \lambda_2\}$. *We also have the following productions*

$$\begin{aligned} \lambda_1 &\implies \mid \mu_1 \mid \mu_2, \\ \lambda_2 &\implies \mid \mu_3 \mid \mu_4, \end{aligned} \quad (1)$$

*where* $\mu_1 := \lambda_2 a \lambda_2$, $\mu_2 := \lambda_2 B \lambda_2$, $\mu_3 := CcA$, *and* $\mu_4 := aBb$. *Using the productions in* (1), *we can generate the following strings:* $CcAaCcA$, $aBbaaBb$, $aBbaCcA$, $aBbBaBb$, *etc. Note that in a CFG, we can also consider a single line of production or a subgroup of the productions to generate strings. For example, using the second line in* (1), *we generate* $Cca$, $aBb$.

### D. Formulation of Integer Encoding of LTL Formulas

Next, we present the integer encoding of LTL formulas in [64], which will be used when solving the model discrimination problem in the Section VI.

The time evaluation of a given formula $\phi$ is given by $\boldsymbol{P}_\phi \triangleq P_\phi^0 P_\phi^1 \ldots P_\phi^T$. For brevity, we will now only present constraints for the satisfaction of each operator of the LTL semantics, i.e., $(\boldsymbol{\sigma}, t) \models \phi$ for the following operators, where $\pi, \pi'$ and $\pi_i$ are atomic propositions, and $P_\phi^t$ is the truth value of formula $\phi$ at time $t$, as defined in [64].

*Negation:* The formula $\phi = \neg \pi$ can be modeled as:

$$P_\phi^t = (1 - P_\pi^t). \quad (2)$$

*Disjunction:* The formula $\phi = \bigvee_{i=1}^k \pi_i$ can be modeled as:

$$P_\phi^t \leq \Sigma_{i=1}^k P_{\pi_i}^t; \quad P_\phi^t \geq P_{\pi_i}^t, i \in \mathbb{Z}_1^k. \quad (3)$$

*Conjunction:* The formula $\phi = \bigwedge_{i=1}^k \pi_i$ can be modeled as:

$$P_\phi^t \geq \Sigma_{i=1}^k P_{\pi_i}^t - (k-1); \quad P_\phi^t \leq P_{\pi_i}^t, i \in \mathbb{Z}_1^k. \quad (4)$$

*Next:* The formula $\phi = \mathbf{X} p$ can be modeled as:

$$P_\phi^t = P_\pi^{t+1}. \quad (5)$$

*Until:* The formula $\phi = \pi' \, \mathbf{U} \, \pi$ can be modeled as:

$$\begin{aligned} \alpha_{tj} &\geq P_\pi^j + \Sigma_{\tau=t}^{j-1} P_{\pi'}^\tau - (j-t+1), j \in \mathbb{Z}_{t+1}^T; \\ \alpha_{tj} &\leq P_\pi^j, \, \alpha_{tj} \leq P_{\pi'}^\tau, j \in \mathbb{Z}_{t+1}^T, \tau \in \mathbb{Z}_t^j; \\ P_\phi^t &\leq \Sigma_{j=t}^T \alpha_{tj}; \\ P_\phi^t &\geq \alpha_{tj}, j \in \mathbb{Z}_t^T. \end{aligned} \quad (6)$$

*Eventually:* The formula $\phi = \mathbf{F} \pi$ can be modeled as:

$$P_\phi^t \leq \Sigma_{\tau=t}^T P_\pi^\tau; \quad P_\phi^t \geq P_\pi^\tau, \tau \in \mathbb{Z}_t^T. \quad (7)$$

*Always:* The formula $\phi = \mathbf{G} \pi$ can be modeled as:

$$P_\phi^t \geq \Sigma_{\tau=t}^T P_\pi^\tau - (T-t); \quad P_\phi^t \leq P_\pi^\tau, \tau \in \mathbb{Z}_t^T. \quad (8)$$

## III. PROBLEM FORMULATION

### A. Modeling Framework

In this paper, we consider two types of atomic propositions. One type of propositions represents modes of the system dynamics and the other type represents constraint modes indicating if certain state constraints are active or inactive; thus, we consider $\Sigma = \Sigma_m \times \Sigma_s$ as a finite set of atomic propositions with $\Sigma_m \subseteq \Sigma$ being the set of switched system modes and $\Sigma_s \subseteq \Sigma$ being the set of state-dependent constraint modes. Specifically, we consider the $N_p$ different

unknown model-task pair $(\{\mathcal{G}^l\}_{l=1}^{N_p}, \phi^l)$, given by:

$$x_{t+1} = \begin{cases} f^l_{\sigma^{m,1}}(x_t, u_t) + w_t, & \text{if } \sigma^m_t = e_1, \\ \vdots \\ f^l_{\sigma^{m,|\Sigma_m|}}(x_t, u_t) + w_t, & \text{if } \sigma^m_t = e_{|\Sigma_m|}, \end{cases} \quad (9)$$
$$x^c_{t,j} = g^l_{\sigma^{s,j}}(z_{t,j}), \text{ if } \sigma^{s,j}_t = 1, \ \forall j \in \mathbb{Z}^{|\Sigma_s|}_1,$$
$$(\boldsymbol{\sigma}, t) \models \phi^l,$$
$$y_t = Cx_t + v_t,$$

where $x_t \in \mathcal{X} \subseteq \mathbb{R}^{n_x}$ is the state vector at time $t \in \mathbb{N}$, $u_t \in \mathcal{U} \subseteq \mathbb{R}^{n_u}$ is the control input vector, $y_t \in \mathcal{Y} \subseteq \mathbb{R}^{n_y}$ is the observed output vector, $w_t \in \mathcal{W} \subseteq \mathbb{R}^{n_w}$ is the bounded process noise and $v_t \in \mathcal{V} \subseteq \mathbb{R}^{n_v}$ is the bounded measurement noise, where we assume that the sets $\mathcal{X}, \mathcal{U}, \mathcal{W}, \mathcal{Y},$ and $\mathcal{V}$ are known polytopes. $e_i$ is the unit basis vector, $\sigma_t \triangleq [\sigma^{m\top}_t \ \sigma^{s\top}_t]^\top \in \{0,1\}^{|\Sigma_m|+|\Sigma_s|}$ and $f^l_{\sigma^{m,i}_t} : \mathcal{X} \times \mathcal{U} \to \mathbb{R}^n$ are vector fields describing the unknown state dynamics for each mode represented by atomic proposition $\sigma^m_t = e_i$, while $g^l_{\sigma^{s,j}_t} : \mathcal{X}^c_j \to \mathcal{Z}_j$ are vector fields describing the unknown state-dependent modes for state constraints corresponding to each atomic proposition $\sigma^{s,j}_t = 1$. For each $j$, $x^c_{t,j}$ and $z_{t,j}$ are disjoint subsets of $x_t$ (such that we do not have implicit equations in (9)) and correspondingly, their domains $\mathcal{X}^c_j$ and $\mathcal{Z}_j$ are also disjoint. The trace $\boldsymbol{\sigma}$ satisfies an LTL formula $\phi^l$ whose specification is unknown.

Moreover, when the exact state of the unknown systems are also unknown (a common scenario), we can adopt auto-regressive models for (9), i.e., $x_t$ can be selected as $[y^\top_t, y^\top_{t-1}, \ldots y^\top_{t+1-n_{xy}}]^\top$, which is concatenation of the measured output $y$ for the current and previous $n_{xy}$ time steps, $u_t$ can similarly be a concatenation of system inputs from the current and previous time steps, and the $C$ matrix in (9) can be chosen as $[I_{n_y \times n_y} \ \mathbf{0}]$. Further, we assume that the unknown vector fields are Lipschitz continuous.

**Assumption 1.** *Each unknown vector field $f(\cdot)$ (or $g(\cdot)$), is $L_p$-Lipschitz continuous, i.e., there exists a positive finite-valued $L_p > 0$, called the Lipschitz constant, such that for all $x_1, x_2$ in the domain of $f$, $|f(x_2) - f(x_1)| \leq L^f_p \|x_2 - x_1\|_p$ for $p \in \{1, \infty\}$ and similarly, for all $z_{1,i}, z_{2,i}$ in the domain of $g$, $|g(z_{2,i}) - g(z_{1,i})| \leq L^f_p \|z_{2,i} - z_{1,i}\|_p$ for $p \in \{1, \infty\}$.*

In addition, we assume that we only have access to traces data set $\mathcal{D}^l_{LTL}$, as well as the input-output trajectories data set $\mathcal{D}^l_\sigma$ for each $\sigma \in \Sigma$.

*B. Problem Statement*

To formulate the problems of interest, we adopt the following definitions.

**Definition 1.** *We define a length-$T$ trace as a finite word $\boldsymbol{\sigma}_T = \sigma_{t_0}\sigma_{t_0+1}\ldots\sigma_{t_0+T-1}$, where $\sigma_t \in \Sigma$ and $T$ is a finite time horizon. In other words, $\boldsymbol{\sigma}_T$ is the length-$T$ prefix of $\boldsymbol{\sigma}$.*

**Definition 2.** *We define a length-$T$ trajectory as $\eta_T = \{u_t, \sigma_t, y_t\}^{t_0+T-1}_{t=t_0}$, where $u_t \in \mathcal{U}$ is the control input and $\mathcal{U} \subseteq \mathbb{R}^m$, $\sigma_t \in \Sigma$ is the mode at time $t$ from the finite set of modes $\Sigma$, $y_t$ is the system output, and $T$ is a finite time*

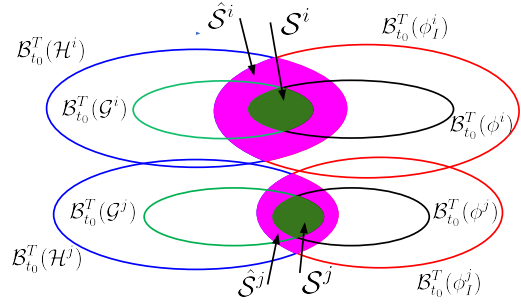

Fig. 2: For two model-task pairs $(\mathcal{G}^i, \phi^i)$ and $(\mathcal{G}^j, \phi^j)$ where the true system dynamics model and LTL specification of both pairs are unknown, we aim to find the $\mathcal{H}$ and $\phi_I$ from data so that the original system behavior is included by inferred system behavior ($\mathcal{S} \subseteq \hat{\mathcal{S}}$). Therefore, if the bigger magenta sets for model-task pairs $i$ and $j$ do not intersect, the two unknown model-task pairs are also discriminated.

*horizon.*

**Definition 3.** *We define the length-$T$ behavior of a model $\mathcal{G}^l$ at time $t_0$ as the set of all length-$T$ trajectories $\eta_T$ that are compatible with $\mathcal{G}^l$ denoted by $\mathcal{B}^T_{t_0}(\mathcal{G}^l)$.*

**Definition 4.** *We define the length-$T$ behavior of a specification $\phi^l$ at time $t_0$ as a set of all length-$T$ traces $\boldsymbol{\sigma}_T$ that satisfy $\phi^l$ (i.e., $(\boldsymbol{\sigma}, t_0) \models \phi^l$) denoted by $\mathcal{B}^T_{t_0}(\phi^l)$.*

Note in Definition 3, the mode sequences are not constrained by any specifications/LTL formulas. Further, the length-$T$ behaviors of the to-be-learned set-membership model $\mathcal{H}^l$ and the to-be-inferred specification $\phi^l_I$ can be similarly defined.

Having defined the above, we can now state our first problem of learning the unknown switched system dynamics $\mathcal{G}^l$ and inferring the unknown LTL specification $\phi^l$ from data with some specific inclusion properties:

**Problem 1** (Model Inference/Learning). *Given (prior) data sets $\mathcal{D}^l_{LTL}$ and $\mathcal{D}^l_\sigma$ for each $\sigma \in \Sigma$ that are generated by the unknown model-task (i.e., dynamics-specification) pair $(\mathcal{G}^l, \phi^l)$ for all $l \in \mathbb{Z}^{N_p}_1$, learn an over-approximation model $\mathcal{H}^l$ of the original model $\mathcal{G}^l$, s.t., $\mathcal{B}^T_{t_0}(\mathcal{G}^l) \subseteq \mathcal{B}^T_{t_0}(\mathcal{H}^l)$ for all $t_0$ and $T$, and infer the inferred specification $\phi^l_I$ such that $\phi^l \to \phi^l_I$ (and equivalently, $\mathcal{B}^T_{t_0}(\phi^l) \subseteq \mathcal{B}^T_{t_0}(\phi^l_I)$ for all $t_0$ and $T$).*

Given the design that $\mathcal{H}^l$ is an over-approximation of $\mathcal{G}^l$ and $\phi^l_I$ is implied by $\phi^l$ as well as the definitions of system behaviors, we show that the intersection of the bigger circles (cf. Fig. 2) represent all possible learned models and inferred LTL specifications that are consistent with the data (including generalization errors) in following proposition:

**Proposition 1.** *For the set $\mathcal{S}^l = \mathcal{B}^T_{t_0}(\phi^l) \cap \mathcal{B}^T_{t_0}(\mathcal{G}^l)$ and the set $\hat{\mathcal{S}}^l = \mathcal{B}^T_{t_0}(\phi^l_I) \cap \mathcal{B}^T_{t_0}(\mathcal{H}^l)$ containing the length-$T$ sequences generated by the true model-task pair $(\mathcal{G}^l, \phi^l)$ and learned/inferred model-task pair $(\mathcal{H}^l, \phi^l_I)$, respectively, if $\mathcal{B}^T_{t_0}(\mathcal{G}^l) \subseteq \mathcal{B}^T_{t_0}(\mathcal{H}^l)$ and $\phi^l \to \phi^l_I$ (i.e., $\mathcal{B}^T_{t_0}(\phi^l) \subseteq \mathcal{B}^T_{t_0}(\phi^l_I)$), then we have $\mathcal{S}^l \subseteq \hat{\mathcal{S}}^l$.*

*Proof.* We assume that $\eta \in \mathcal{S}^l = \mathcal{B}_{t_0}^T(\phi^l) \cap \mathcal{B}_{t_0}^T(\mathcal{G}^l)$ is an arbitrary length-$T$ sequence generated by true model-task pair $(\mathcal{G}^l, \phi^l)$: 1) for any $\eta \in \mathcal{S}^l$, we have $(\eta, t_0) \models \phi^l$; thus, if $\phi^l \to \phi_I^l$, then we conclude that for any $\eta \in \mathcal{S}^l$ we have $(\eta, t_0) \models \phi_I^l$ (i.e., for any $\eta \in \mathcal{S}^l$, we have $\eta \in \mathcal{B}_{t_0}^T(\phi_I^l)$); 2) we know that $\eta \in \mathcal{B}_{t_0}^T(\mathcal{G}^l)$ and $\mathcal{B}_{t_0}^T(\mathcal{G}^l) \subseteq \mathcal{B}_{t_0}^T(\mathcal{H}^l)$; therefore, for any $\eta \in \mathcal{S}^l$, we have $\eta \in \mathcal{B}_{t_0}^T(\mathcal{H}^l)$. By considering 1) and 2), we conclude that $\mathcal{S} \subseteq \hat{\mathcal{S}}$, where $\hat{\mathcal{S}}^l = \mathcal{B}_{t_0}^T(\phi_I^l) \cap \mathcal{B}_{t_0}^T(\mathcal{H}^l)$. □

Then, if the bigger magenta sets for model-task pairs $i$ and $j$ do not intersect (cf. Fig. 2), i.e., $\hat{\mathcal{S}}^i \cap \hat{\mathcal{S}}^i = \emptyset$, it is straightforward to conclude that the behaviors of original model-task pair (green sets) are also disjoint:

**Proposition 2.** *For the true model-task pairs $(\mathcal{G}^i, \phi^i)$ and $(\mathcal{G}^j, \phi^j)$, and the learned/inferred model-task pairs $(\mathcal{H}^i, \phi_I^i)$ and $(\mathcal{H}^j, \phi_I^j)$, we have $\hat{\mathcal{S}}^i \cap \hat{\mathcal{S}}^j = \emptyset \to \mathcal{S}^i \cap \mathcal{S}^j = \emptyset$, where $\hat{\mathcal{S}}^i = \mathcal{B}_{t_0}^T(\phi_I^i) \cap \mathcal{B}_{t_0}^T(\mathcal{H}^i)$, $\hat{\mathcal{S}}^j = \mathcal{B}_{t_0}^T(\phi_I^j) \cap \mathcal{B}_{t_0}^T(\mathcal{H}^j)$, $\mathcal{S}^i = \mathcal{B}_{t_0}^T(\phi^i) \cap \mathcal{B}_{t_0}^T(\mathcal{G}^i)$, and $\mathcal{S}^j = \mathcal{B}_{t_0}^T(\phi^j) \cap \mathcal{B}_{t_0}^T(\mathcal{G}^j)$.*

Equipped by Propositions 1 and 2, we now state the following data-driven model discrimination problems to determine how long it takes for the model-task pairs in a model-task pair set to be distinguished from each other by way of the distinction of their learned/inferred model-task pairs:

**Problem 2.** *(Distinguishability for a set of true model-task pairs). Given a set of learned/inferred model-task pairs $\{(\mathcal{H}^l, \phi_I^l)\}_{l=1}^{N_p}$ and a finite time horizon $T$, determine whether the set of learned/inferred model-task pairs is $T$-distinguishable, i.e., whether $\exists t_0$ such that $\bigcap_{l=1}^{N_p}(\mathcal{B}_{t_0}^T(\mathcal{H}^l) \cap \mathcal{B}_{t_0}^T(\phi_I^l)) = \emptyset$.*

**Problem 3.** *(Model-task pair discrimination) Given a (newly observed) length-$T$ input-mode-output sequence $\{u_t, \sigma_t, y_t\}_{t=t_c-T+1}^{t_c}$ at the current time $t_c$ and a set of learned/inferred model-task pairs $\{(\mathcal{H}^l, \phi_I^l)\}_{l=1}^{N_p}$, determine which learned/inferred model-task pair it is based on the given sequence. That is to find an $i \in \mathbb{Z}_1^{N_p} = \{1, 2, ..., N_p\}$ that satisfies $(\mathcal{B}_{t_0}^T(\mathcal{H}^i) \cap \mathcal{B}_{t_0}^T(\phi_I^i) \neq \emptyset) \wedge (\mathcal{B}_{t_0}^T(\mathcal{H}^j) \cap \mathcal{B}_{t_0}^T(\phi_I^j) = \emptyset)$, $\forall j \in \mathbb{Z}_1^{N_p}$ and $i \neq j$ with $t_0 = t_c - T + 1$.*

Our approach to solve Problem 1 consists of a few steps. We first divide the data into trajectory and trace data, and then, we infer the system dynamics and LTL specifications separately in Sections IV and V, respectively. In addition, when prior information about the family of possible LTL specifications is known or given, we also consider formula reduction techniques in Section V-V-B. Finally, using the learned system dynamics and inferred LTL formulas (from Problem 1), we address Problems 2 and 3 in Section VI.

## IV. SYSTEM DYNAMICS LEARNING

In this section, we learn over-approximations of the unknown true system dynamics from the trajectory data $\mathcal{D}_\pi^l$ for all $\pi \in \Sigma$ using set-membership methods in [35], [65], [66]. Given an arbitrary original (possibly unknown) vector field/function $r_o = f_a(r) : \mathbb{R} \subset \mathbb{R}^{n_r} \to \mathbb{R}$, the goal of an over-approximation procedure is to find a pair of functions $\underline{f}_a$ and $\overline{f}_a$ (i.e., to find an over-approximation model $\mathcal{H} \triangleq \{\overline{f}_a, \underline{f}_a\}$) such that the function $f_a(\cdot)$ is bounded by the pair of functions, i.e., $\underline{f}_a$ and $\overline{f}_a$ satisfy the following:

$$\underline{f}_a(r) \leq f_a(r) \leq \overline{f}_a(r), \ \forall r \in \mathbb{R}.$$

If only a noisy sampled input-output data set $\mathcal{D}_r = \{(\tilde{r}_j, \tilde{r}_{o,j}) | j = 1, \ldots, N_D\}$ is available and function $f_a(\cdot)$ is assumed to be Lipschitz continuous, a data abstraction can be found by following proposition according to [35].

**Proposition 3.** *Consider a unknown Lipschitz function $f_a(\cdot)$ and its corresponding noisy data set $\mathcal{D}_r = \{(\tilde{r}_j, \tilde{r}_{o,j}) | j = 1, \ldots, N_D\}$. For all $r \in \mathbb{R}$, $\underline{f}_{a,\mathcal{D}}(\cdot)$ and $\overline{f}_{a,\mathcal{D}}(\cdot)$ are lower and upper abstraction functions for unknown function $f_a(\cdot)$, i.e., $\forall r \in \mathbb{R}$, $\underline{f}_{a,\mathcal{D}}(r) \leq f_a(r) \leq \overline{f}_{a,\mathcal{D}}(r)$,*

$$\overline{f}_{a,\mathcal{D}}(r) = \min_{j \in \{1,\ldots,N_D\}} (\tilde{r}_{o,j} + L_p \|r - \tilde{r}_j\|_p) + \varepsilon_t,$$
$$\underline{f}_{a,\mathcal{D}}(r) = \max_{j \in \{1,\ldots,N_D\}} (\tilde{r}_{o,j} - L_p \|r - \tilde{r}_j\|_p) - \varepsilon_t,$$

*with a selected norm $p \in \{0, 1, \infty\}$ and $\varepsilon_t \triangleq \varepsilon_{r_o} + (L_p + 1)\varepsilon_r$, where $\varepsilon_r \triangleq \max(\tilde{r} - r)$ and $\varepsilon_{r_o} \triangleq \max(\tilde{r}_o - r_o)$ are bounds on the measurement noise/errors for $r$ and $r_o$, respectively.*

When the Lipschitz constant is unknown, the Lipschitz constant can also be estimated from the dataset $\mathcal{D}_r = \{(\tilde{r}_j, \tilde{r}_{o,j}) | j = 1, \ldots, N_D\}$ as follows, according to [35]:

$$\hat{L}_p^{f_a} = \max_{j \neq i} \frac{|\tilde{r}_{o,j} - \tilde{r}_{oi}| - 2\varepsilon_{r_o}}{\|\tilde{r}_j - \tilde{r}_i\|_p + 2\varepsilon_r}.$$

The method can be applied to find an inclusion model that over-approximates the unknown system dynamics. However, this data-driven approach belongs to the class of non-parametric learning methods in the machine learning literature that are known to not scale well with the size of the data sets. Thus, to further simplify the data-driven model in $\mathcal{H}^l$, i.e., the functions $f_{\sigma^m}^l$ and $g_{\sigma^{s,j}}^l$ for each $\sigma^m \in \{e_1, \ldots, e_{|\Sigma_m|}\}$ and each $\sigma^s \in \{0, 1\}^{|\Sigma_s|}$, we propose to leverage a result in [66], [67] to over-approximate the nonlinearities with piecewise affine inclusions. First, we introduce the definition of the partitions.

**Definition 5** (Partition). *For each function $f_{\sigma^m}^l$ or $g_{\sigma^{s,j}}^l$, a partition $\mathcal{I}_\sigma^{f,l}$ of the closed bounded region $\mathcal{X} \times \mathcal{U} \subseteq \mathbb{R}^{n_x+n_u}$ is a collection of $q_{\sigma^m}^{f,l}$ subregions $\mathcal{I}_{\sigma^m}^{f,l} = \{I_{\sigma^m,\star}^{f,l} \mid \star \in \mathbb{Z}_1^{q_{\sigma^m}^{f,l}}\}$ such that $\mathcal{X} \times \mathcal{U} \subseteq \bigcup_{\star=1}^{q_{\sigma^m}^{f,l}} I_{\sigma^m,\star}^{f,l}$ and $I_{\sigma^m,\star}^{f,l} \cap I_{\sigma^m,\star'}^{f,l} = \partial I_{\sigma^m,\star}^{f,l} \cap \partial I_{\sigma^m,\star'}^{f,l}$, $\forall \star \neq \star' \in \mathbb{Z}_1^q$, where $\partial I_{\sigma^m,\star}^{f,l}$ is the boundary of set $I_{\sigma^m,\star}^{f,l}$. Similarly, a partition $\mathcal{I}_{\sigma^{s,j}}^{g,l} = \{I_{\sigma^{s,j},\dagger}^{g,l} \mid \dagger \in \mathbb{Z}_1^{q_{\sigma^{s,j}}^{g,l}}\}$ with $q_{\sigma^{s,j}}^{g,l}$ subregions of the closed bounded region $\mathcal{Z}_j$ for vector fields $g_{\sigma^{s,j}}^l$ can be defined.*

We assume that the partitions are polytopic. Then, for each polytopic subregion $I_{\sigma^m,\star}^{f,l} \in \mathcal{I}_{\sigma^m}^{f,l}$ (or $I_{\sigma^{s,j},\dagger}^{g,l} \in \mathcal{I}_{\sigma^{s,j}}^{g,l}$) that partitions the domain of interest, the unknown function $f_{\sigma^m}^l$ (or $g_{\sigma^{s,j}}^l$) can be over-approximated/abstracted by a pair of affine functions $\underline{f}^l_{\sigma^m,\star}$, $\overline{f}^l_{\sigma^m,\star}$ (or $\underline{g}^l_{\sigma^{s,j},\dagger}$, $\overline{g}^l_{\sigma^{s,j},\dagger}$) by solving a linear programming (LP) problem similar to [66], [67]. As a result, for all $(x,u) \in I^{f,l}_{\sigma^m,\star}$ (or $z_j \in I^{g,l}_{\sigma^{s,j},\dagger}$), the function $f^l_{\sigma^m}(x,u)$ (or $g^l_{\sigma^{s,j}}(z_j)$) is sandwiched/framed by a pair of affine functions, i.e., $\underline{f}^l_{\sigma^m}(x,u) \leq \underline{f}^l_{D,\sigma^m}(x,u) \leq \overline{f}^l_{D,\sigma^m}(x,u) \leq \overline{f}^l_{\sigma^m}(x,u)$ (or $\underline{g}^l_{\sigma^{s,j}}(z_j) \leq \underline{g}^l_{D,\sigma^{s,j}}(z_j) \leq \overline{g}^l_{D,\sigma^{s,j}_j}(z_j) \leq \overline{g}^l_{\sigma^{s,j}}(z_j)$) with

$$\underline{f}^l_{\sigma^m,\star}(x,u) = \underline{A}^l_{\sigma^m,\star} x + \underline{B}^l_{\sigma^m,\star} + \underline{h}^{f,l}_{\sigma^m,\star},$$
$$\overline{f}^l_{\sigma^m,\star}(x,u) = \overline{A}^l_{\sigma^m,\star} x + \overline{B}^l_{\sigma^m,\star} u + \overline{h}^{f,l}_{\sigma^m,\star},$$
$$\underline{g}^l_{\sigma^{s,j},\dagger}(z_j) = \underline{H}^l_{\sigma^{s,j},\dagger} z_j + \underline{h}^{g,l}_{\sigma^{s,j},\dagger},$$
$$\overline{g}^l_{\sigma^{s,j},\dagger}(z_j) = \overline{H}^l_{\sigma^{s,j},\dagger} z_j + \overline{h}^{g,l}_{\sigma^{s,j},\dagger},$$

where $\underline{A}^l_{\sigma^m,\star}$, $\overline{A}^l_{\sigma^m,\star}$, $\underline{B}^l_{\sigma^m,\star}$, $\overline{B}^l_{\sigma^m,\star}$, $\underline{H}^l_{\sigma^{s,j},\dagger}$, $\overline{H}^l_{\sigma^{s,j},\dagger}$, $\underline{h}^{f,l}_{\sigma^m,\star}$, $\overline{h}^{f,l}_{\sigma^m,\star}$, $\underline{h}^{g,l}_{\sigma^{s,j},\dagger}$ and $\overline{h}^{g,l}_{\sigma^{s,j},\dagger}$ are of appropriate dimensions and are constants that are determined by the piecewise affine abstraction algorithm in [66], [67].

Since the original functions $f(\cdot)$ and $g(\cdot)$ are bounded by data-driven over-approximation $\overline{f}_D(\cdot)$ and $\underline{f}_D(\cdot)$, the following piecewise affine interval models $\mathcal{H}^l$ is an over-approximation for the original dynamics model $\mathcal{G}^l$ satisfying $\mathcal{B}^T_{t_0}(\mathcal{G}^l) \subseteq \mathcal{B}^T_{t_0}(\mathcal{H}^l)$ for all $t_0$ and $T$:

$$\begin{pmatrix} \underline{A}^l_{\sigma^m_t,\star} x_t + \underline{B}^l_{\sigma^m_t,\star} u_t \\ +\underline{h}^{f,l}_{\sigma^m_t,\star} \end{pmatrix} \leq x_{t+1} \leq \begin{pmatrix} \overline{A}^l_{\sigma^m_t,\star} x_t + \overline{B}^l_{\sigma^m_t,\star} u_t \\ +\overline{h}^{f,l}_{\sigma^m_{t,i},\star} \end{pmatrix},$$
$$y_t = C x_t + v_t,$$
$$\underline{H}^l_{\sigma^{s,j}_t,\dagger} z_{t,j} + \underline{h}^{g,l}_{\sigma^{s,j}_t,\dagger} \leq x^c_{t,j} \leq \overline{H}^l_{\sigma^{s,j}_t,\dagger} z_{t,j} + \overline{h}^{g,l}_{\sigma^{s,j}_t,\dagger},$$

where their corresponding polytopic subregions $I^{f,l}_{\sigma^m_{t,i},\star}$ and $I^{g,l}_{\sigma^m_{t,j},\dagger}$ can be represented by the following linear constraints:

$$S^{x,l}_{\sigma^m_t,\star} x_t + S^{u,l}_{\sigma^m_t,\star} u_t \leq \beta^l_{\sigma^m_t,\star},$$
$$S^{z,l}_{\sigma^{s,j}_t,\dagger} z_t \leq \beta^l_{\sigma^{s,j}_t,\dagger},$$

respectively, with $S^{x,l}_{\sigma^m_t,\star}$, $S^{u,l}_{\sigma^m_t,\star}$, $S^{z,l}_{\sigma^{s,j}_t,\dagger}$, $\beta^l_{\sigma^m_t,\star}$ and $\beta^l_{\sigma^{s,j}_t,\dagger}$ of appropriate dimensions.

Note that the precision of the inferred/learned model can be improved with more and better chosen *partitions*, but may result in longer computation times since more integer variables will be introduced in our solutions in Section VI.

## V. LTL SPECIFICATION INFERENCE

We first propose an algorithm for learning the inferred specification $\phi_I$ (without any prior information). Then, we propose a framework for reducing the size of the inferred formula, when prior information is available or given, for further improving the computational efficiency of the model discrimination algorithms in Section VI.

### A. Learning the Unknown LTL Specification

In this subsection, we present an algorithm for learning $\phi_I$ from a given data set of traces $\mathcal{D}_{LTL}$. Algorithm 1 illustrates the framework that we use to learn $\phi_I$ where the upper bound on the size of the true LTL formula, denoted by $Itr$, is given.

---

**Algorithm 1:** Inference of LTL Formulas

**Input:** $\mathcal{D}_{LTL} \in \Sigma$
Maximum formula size $Itr \in \mathbb{N}$

1 $E_\gamma = E_{LTL} = E' = \emptyset$
2 $i \leftarrow 1$
3 $k \leftarrow 1$
4 **while** $k \leq Itr$ **do**
5    Construct $\Phi^{\mathcal{D}_{LTL},\gamma}_k$ of size $k$ consistent with $\mathcal{D}_{LTL}$
6    **if** $\Phi^{\mathcal{D}_{LTL},\gamma}_k$ *is satisfiable* **then**
7      Compute a model $\gamma$ of $\Phi^{\mathcal{D}_{LTL},\gamma}_k$
8      **while** *a distinct LTL formula using $\gamma$ can be constructed* **do**
9        Construct LTL formula $\phi^i$ with model $\gamma$
10        Store $\phi^i$ in $E'$ and in $E_{LTL}$
11        Add $\bigwedge_{j=1,\ldots,|E'|-1} \bigvee_{n_l \in \mathcal{E}^j} (n_l \notin \mathcal{E}^j)$ to the constraints $\Phi^{\mathcal{D}_{LTL},\gamma}_k$
12        $i \leftarrow i+1$
13      **end while**
14      $E' = \emptyset$
15      Store $\gamma$ in $E_\gamma$
16      Add $\bigwedge_{k'=1,\ldots,|E_\gamma|-1} \bigvee_{\rho \in \gamma_{k'}} (\rho \neq \gamma_{k'}(\rho))$ to the constraints $\Phi^{\mathcal{D}_{LTL},\gamma}_k$
17    **else**
18      $E_\gamma = \emptyset$
19      $k \leftarrow k+1$
20    **end if**
21 **end while**
22 $N_o \leftarrow i-1$
23 $\phi_I \leftarrow \bigvee_{i=1,\ldots,|E_{LTL}|} \phi^i$
24 **return** $\phi_I$

---

In Algorithm 1, we convert the problem of inferring LTL formulas from $\mathcal{D}_{LTL}$ to a satisfiability problem in the propositional logic. The satisfiabilty problem is about assessing whether a logical formula is satisfiable or not. For the satisfiabilty problem in the Boolean domain, we can use *Boolean satisfiability* (SAT) solvers. One of the common SAT solvers is the *Z3 theorem prover* [68].

We use propositional formulas in the satisfiability problem to infer LTL formulas. If $\mathcal{P}$ is a set of propositional variables in the Boolean domain, then a propositional variable $\rho \in \mathcal{P}$ is a propositional formula. In addition, if $\Phi$ and $\Psi$ are propositional formulas, then $\neg \Phi$ and $\Phi \vee \Psi$ are propositional formulas as well. Then, we define a model[1] of a propositional formula as a mapping $\gamma : \mathcal{P} \to \mathbb{B}$. The semantics of this propositional valuation is given by a satisfaction relation that is defined as follows. $\rho \models \gamma$ if and only if $\gamma(\rho) = 1$, $\gamma \models \neg \Phi$ if and only if $\gamma \not\models \Phi$, $\gamma \models \Phi \vee \Psi$ if and only if $\gamma \models \Phi$ or $\gamma \models \Psi$, and finally, if $\gamma \models \Phi$, then we introduce $\gamma$ as a *model* of $\Phi$. In the case that such a model $\gamma$ exists, the propositional formula $\Phi$ is satisfiable. Such a model provides us with sufficient information to construct an LTL formula from that model. To infer LTL formulas using propositional formulas, we use the framework proposed in [59]. Using this framework, we can construct a propositional formula of a given size that is consistent with the data set of traces $\mathcal{D}_{LTL}$.

---

[1]The term "model" in Subsection V refers to a model of a proportional formula and is different from the system model used in the rest of the paper.

By exploiting the fact that for a propositional formula, multiple distinguished models can be found, we can construct multiple distinct LTL formulas from each model. Hence, the key idea in Algorithm 1 is to infer all possible LTL formulas with the maximum size of $Itr$ from $\mathcal{D}_{LTL}$ using propositional formulas.

Algorithm 1 starts from $k = 1$ and constructs a propositional formula $\Phi_k^{\mathcal{D}_{LTL},\gamma}$ consistent with $\mathcal{D}_{LTL}$ where $k$ is the size of the propositional formula (Line 5 in Alg. 1). Then Algorithm 1 computes a model $\gamma$ of $\Phi_k^{\mathcal{D}_{LTL},\gamma}$, if $\Phi_k^{\mathcal{D}_{LTL},\gamma}$ is satisfiable (Line 7 in Alg. 1). After that, we construct all the possible LTL formulas from the computed model $\gamma$ (Lines 8 to 13 in Alg. 1). In Algorithm 1, for inferring each LTL formula, we first construct a DAG from the model $\gamma$ and then label each node accordingly (such that the corresponding LTL formula is consistent with $\mathcal{D}_{LTL}$) [59]. The constraint in Line 11 of Algorithm 1 ensures that the already constructed LTL formula from $\gamma$ will not be constructed again. Moreover, the constraint in Line 16 ensures that the already computed model will not be computed again.

If $\Phi_k^{\mathcal{D}_{LTL},\gamma}$ of size $k$ is not satisfiable, then $k$ increases by 1 to infer LTL formulas using the propositional formula $\Phi_{k+1}^{\mathcal{D}_{LTL},\gamma}$ of size $k+1$ (Lines 17 to 20 in Alg. 1). The outer *while loop* in Algorithm 1 terminates once $N_o$ possible LTL formulas with the maximum size of $Itr$ are inferred from the given data set of traces $\mathcal{D}_{LTL}$. Finally, we construct $\phi_I := \bigvee_{i=1,\ldots,|E_{LTL}|} \phi^i$.

**Theorem 1.** *Given a data set of traces $\mathcal{D}_{LTL}$, if $E_{LTL} \neq \emptyset$, then $\phi \to \phi_I$ holds for $\phi_I := \bigvee_{i=1,\ldots,|E_{LTL}|} \phi^i$ where the size of each $\phi^i \in E_{LTL}$ is at most $Itr$ and $\phi$ is the true LTL formula.*

*Proof.* Algorithm 1 terminates when all the possible LTL formulas with the maximum size of $Itr$ are inferred including the true LTL formula $\phi$, all consistent with the given data set $\mathcal{D}_{LTL}$. Hence, we conclude that if $E_{LTL} \neq \emptyset$, then $\phi \to \phi_I$ holds for $\phi_I := \bigvee_{i=1,\ldots,|E_{LTL}|} \phi^i$ where the size of each $\phi^i \in E_{LTL}$ is at most $Itr$. □

### B. Reducing the Size of the Inferred LTL Formula

In this subsection, we present a novel framework and its associated algorithm for reducing the size of inferred LTL formula $\phi_I$ such that $\phi \to \phi_I$ still holds. The reason is that reducing the size of inferred LTL formula while satisfying $\phi \to \phi_I$ drastically decreases the computational cost in model discrimination. In what follows, we first introduce *prior information* to constrain the size of inferred LTL formula such that $\phi \to \phi_I$. Then, we propose an algorithm to further reduce the size of the inferred formula while satisfying $\phi \to \phi_I$.

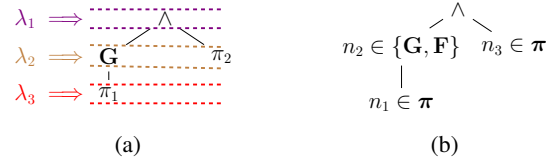

Fig. 3: (a) The DAG corresponding to the LTL formula $(\mathbf{G}\,\pi_1) \wedge \pi_2$ in Example 2. Each layer in the DAG corresponds to a non-terminal $\lambda_i \in \mathcal{N} = \{\lambda_1, \lambda_2, \lambda_3\}$. (b) The DAG structure representing a given $\mathcal{PI}$. $CFG_{\mathcal{PI}}$ corresponding to the given $\mathcal{PI}$ is defined in (12) with $\boldsymbol{\pi} \triangleq \{\pi_1, \pi_2, \pi_1\}$.

*1) Prior Information*

Here, we introduce *prior information* that we use for reducing the size of inferred LTL formula $\phi_I$, when it is available/given. Before introducing *prior information*, we need to explain how we use CFG to generate strings of LTL formulas according to a given syntax DAG.

As mentioned earlier, an LTL formula can be represented as a syntax DAG. This DAG can be encoded as a CFG. In defining the productions in CFG corresponding to a DAG, the head of the first production corresponds to the root node and the head of the last production corresponds to the outermost child node in a DAG.

**Example 2.** *The DAG corresponding to the LTL formula $(\mathbf{G}\,\pi_1) \wedge \pi_2$ is shown in Figure 3a with the corresponding CFG defined in (11), where $\lambda_1$ corresponds to the root node $\wedge$ and $\lambda_3$ corresponds to the outermost child node $\pi_1$. Also, each layer in the DAG corresponds to a non-terminal $\lambda_i \in \mathcal{N} = \{\lambda_1, \lambda_2, \lambda_3\}$ and $\Pi = \{\mathbf{G}, \pi_1, \pi_2\}$.*

$$\begin{aligned} \lambda_1 &\implies \lambda_2 \wedge \lambda_2 \\ \lambda_2 &\implies \mathbf{G}\,\lambda_3 \mid \pi_2 \\ \lambda_3 &\implies \pi_1. \end{aligned} \quad (11)$$

Now, we define *prior information* in the following.

**Definition 6.** *The prior information $\mathcal{PI}$ represents a DAG structure with the following properties.*

1) *The labels of some of the nodes are known and are **fixed**, and the labels of the rest of the nodes are to be inferred which are referred to as **unfixed** nodes.*
2) *The label of each **unfixed** node belongs to a predetermined set of Boolean connectives, temporal operators, and atomic propositions.*
3) *The size of $\mathcal{PI}$ equals $Itr$ which is the given upper bound on the size of the true LTL formula $\phi$.*
4) *The DAG corresponding to the true LTL formula $\phi$ is **consistent with the DAG structure or a partial DAG structure** of $\mathcal{PI}$. We explain the concept of being consistent with the DAG structure or the partial DAG structure of $\mathcal{PI}$ through Example 3. For simplicity, we write an LTL formula $\phi$ is consistent with $\mathcal{PI}$ as an equivalent form of stating that the DAG corresponding to the LTL formula $\phi$ is consistent with the DAG structure or a partial DAG structure of $\mathcal{PI}$.*

Determining the DAG structure of $\mathcal{PI}$, the fixed nodes, and the predetermined sets associated with unfixed nodes depends on the specific scenario at hand. Hence, $\mathcal{PI}$ should be determined according to a certain scenario by a domain expert. For implementing $\mathcal{PI}$, we use CFG which allows us to encode $\mathcal{PI}$ such that we can also infer LTL formulas that are *consistent* with the partial structure of $\mathcal{PI}$. This is specifically beneficial since we do not know the size of the true LTL formula. We use $CFG_{\mathcal{PI}}$ to denote the CFG corresponding to a given $\mathcal{PI}$. Here, for the set of non-terminals, we have $\mathcal{N} = \{\lambda_1, \lambda_1, \ldots, \lambda_{N_k}\}$. For the set of terminals, we have $\Pi \subseteq \Sigma \cup \mathcal{O}$.

**Example 3.** *Fig 3b illustrates the DAG structure representing a given $\mathcal{PI}$. We define the CFG corresponding to the given $\mathcal{PI}$ with the productions defined in* (12). *Here* $\Pi = \{\mathbf{G}, \mathbf{F}, \wedge, \pi_1, \pi_2, \pi_3\}$ *and* $\mathcal{N} = \{\lambda_1, \lambda_2, \lambda_3\}$.

$$\begin{aligned} \lambda_1 &\implies \lambda_2 \wedge \lambda_2 \\ \lambda_2 &\implies \mathbf{G}\,\lambda_3 \mid \mathbf{F}\,\lambda_3 \mid \pi_1 \mid \pi_2 \mid \pi_3 \\ \lambda_3 &\implies \pi_1 \mid \pi_2 \mid \pi_3. \end{aligned} \quad (12)$$

*Possible LTL formulas that can be generated from* (12) *are* $(\mathbf{F}\,\pi_1) \wedge \pi_2$, $(\mathbf{G}\,\pi_1) \wedge \pi_2$, $\pi_1 \wedge \pi_2$, $\mathbf{G}\,\pi_1$, $\mathbf{F}\,\pi_3$, *etc. Note that using CFG for implementing $\mathcal{PI}$ may result in generating LTL formulas that are not consistent with the given $\mathcal{PI}$. For example, the LTL formula $\pi_1 \wedge \pi_2$ is not consistent with $\mathcal{PI}$, since the labels of neither of the children node of $\wedge$ belong to $\{\mathbf{G}, \mathbf{F}\}$. The LTL formula $\mathbf{G}\,\pi_1$ is consistent with partial structure of $\mathcal{PI}$. The LTL formula $(\mathbf{F}\,\pi_1) \wedge \pi_2$ is consistent with the whole DAG structure of $\mathcal{PI}$.*

**Remark 1.** *The set of LTL formulas consistent with $\mathcal{PI}$ is a subset of the LTL formulas that can be generated by $CFG_{\mathcal{PI}}$.*

*2) Constraining the Inference of LTL formulas Using Prior Information*

Here, we explain how we use prior information $\mathcal{PI}$ for reducing the size of $\phi_I$. In the first step, in Algorithm 1, we add $CFG_{\mathcal{PI}}$ as a set of constraints to $\Phi_k^{\mathcal{D}_{LTL},\gamma}$ in Line 5 in Algorithm 1. In this way, if $\Phi_k^{\mathcal{D}_{LTL},\gamma}$ is satisfiable in Line 6 in Algorithm 1, then each computed model $\gamma$ in Line 7 in Algorithm 1 satisfies $CFG_{\mathcal{PI}}$. Consequently, each constructed LTL formula $\phi^i$ in Line 9 in Algorithm 1 is consistent with $CFG_{\mathcal{PI}}$. If any LTL formulas consistent with $CFG_{\mathcal{PI}}$ are inferred, then we store them in the set $E_{CFG}$. Note that $E_{CFG} \subseteq E_{LTL}$.

As explained in Remark 1, it is possible that not all the LTL formulas in $E_{CFG}$ are consistent with $\mathcal{PI}$. Hence, in the second step, if $E_{CFG} \neq \emptyset$, then we keep only those formulas that are consistent with $\mathcal{PI}$ and store them in $E_{\mathcal{PI}}$. Finally, we form $\phi_I := \bigvee_{i=1,\ldots,|E_{\mathcal{PI}}|} \phi^i$ where $\phi^i \in E_{\mathcal{PI}}$.

**Proposition 4.** *Given data set of traces $\mathcal{D}_{LTL}$ and $\mathcal{PI}$, if $E_{CFG} \neq \emptyset$, then $\phi \rightarrow \phi_I$ holds for $\phi_I := \bigvee_{i=1,\ldots,|E_{\mathcal{PI}}|} \phi^i$ where the size of each $\phi^i \in E_{\mathcal{PI}}$ is at most $Itr$ and $\phi$ is the true LTL formula.*

*Proof.* 1) We know that Algorithm 1 terminates when all the possible LTL formulas with the maximum size of $Itr$ are inferred, all consistent with the given data set $\mathcal{D}_{LTL}$ and $CFG_{\mathcal{PI}}$. For $E_{\mathcal{PI}}$, we also have $E_{\mathcal{PI}} \subseteq E_{CFG}$ which implies that the size of all the LTL formulas in $E_{\mathcal{PI}}$ is at most $Itr$ (See Remark 1). 2) If $E_{CFG} \neq \emptyset$, we obtain $E_{\mathcal{PI}}$ which contains all the LTL formulas that satisfy the third property of $\mathcal{PI}$. Note that it is assumed that the third property mentioned in Definition 6 is already satisfied in the given $\mathcal{PI}$.

Using 1) and 2), we conclude that if $E_{CFG} \neq \emptyset$, then $\phi \rightarrow \phi_I$ holds for $\phi_I := \bigvee_{i=1,\ldots,|E_{\mathcal{PI}}|} \phi^i$ where the size of each $\phi^i \in E_{\mathcal{PI}}$ is at most $Itr$. □

*3) Reducing the Number of the Inferred LTL Formulas*

For further increasing computational efficiency (e.g., in model discrimination), we can reduce the size of the inferred formulas in both $E_{LTL}$ and $E_{\mathcal{PI}}$ using Algorithm 2. In Algorithm 2, for each two distinct $\phi^i$ and $\phi^{i'} \in E_{Inf}$, $E_{Inf} \in \{E_{LTL}, E_{\mathcal{PI}}\}$, we check 1) if $\phi^i \rightarrow \phi^{i'}$ and 2) if $\phi^{i'} \rightarrow \phi^i$ (Lines 5 and 6 in Algorithm 2). If 1) holds and 2) does not hold, we keep $\phi^i$ and add it to $E'_{Inf}$. If both 1) and 2) hold and $\phi^{i'}$ is already stored in $E'_{Inf}$, we ignore $\phi^{i'}$ (Lines 6 to 10 in Algorithm 2). Then, we form the set $E_{Red} = E_{Inf}/E'_{Inf}$ using which we obtain $\psi_I := \bigvee_{i=1,\ldots,|E_{Red}|} \phi^i$ where $\phi^i \in E_{Red}$.

**Proposition 5.** *In Algorithm 2, if $E_{Red} \neq \emptyset$, then we obtain $\phi \rightarrow \psi_I := \bigvee_{i=1,\ldots,|E_{Red}|} \phi^i$ where $\phi^i \in E_{Red}$ and $\phi$ is the true LTL formula, and the size of $\psi_I$ is smaller than the size of $\phi_I$.*

*Proof.* In Algorithm 2, $E_{Red}$ contains all the LTL formulas $\phi^i \in E_{Inf}$, $E_{Inf} \in \{E_{LTL}, E_{\mathcal{PI}}\}$ such that $\phi^{i'} \rightarrow \phi^i$ for at least one $\phi^{i'} \in E_{Inf}$ with $i \neq i'$. This implies that $\phi_I \rightarrow \psi_I$ where $\phi_I :=: \bigvee_{i=1,\ldots,|E_{Inf}|} \phi^i$ with $\phi^i \in E_{Inf}$ and $\psi_I := \bigvee_{k=1,\ldots,|E_{Red}|} \phi^k$ with $\phi^k \in E_{Red}$. We also can conclude that $E_{Red} \subseteq E_{Inf}$ and hence $|E_{Red}| \leq |E_{Inf}|$ from which follows that the size of $\psi_I$ is smaller than the size of $\phi_I$.

Moreover, since we already know that $\phi \rightarrow \phi_I$ and $\phi_I \rightarrow \psi_I$, we can write $\phi \rightarrow \psi_I$. □

## VI. $T$-Distinguishability and Model Invalidation

After obtaining the inferred system models and LTL formulas, we then propose a detectability analysis algorithm for $T$-distinguishability (i.e., to solve Problem 2), and the (guaranteed) detection time $T$ is found by solving the problem below with increasing $T$:

**Theorem 2** ($T$-Distinguishability)**.** *A pair of learned piecewise constrained affine inclusion models $\mathcal{H}^i$ and $\mathcal{H}^j$, $i \neq j$, with LTL formulas $\phi_I^i$ and $\phi_I^j$ is $T$-distinguishable if the following is infeasible for any $t_0$:*

$$\text{Find} \quad x_t^\star, v_t^\star, u_t, y_t, s_{t,*}^\star, \tilde{s}_{t,\dagger}^\star, a_{t,*}^\star, \tilde{a}_{t,\dagger}^\star, c_t^{\sigma^m}, b_t^{\sigma^m}$$

$$\text{s.t.} \quad \forall \star \in \{i, j\}, \sigma^m \in \{e_1, \ldots, e_{|\Sigma_m|}\}, * \in \mathbb{Z}_1^{q_{\sigma^m}^{f,\star}},$$

$$\ell \in \mathbb{Z}_1^{|\Sigma_s|}, \sigma^{s,\ell} \in \{0,1\}, \dagger \in \mathbb{Z}_1^{q_{\sigma^{s,\ell}}^{g,\star}}, t \in \mathbb{Z}_{t_0}^{t_0+T-1}:$$

$$x_{t+1}^\star \leq \overline{A}_{\sigma^m,*}^\star x_t^\star + \overline{B}_{\sigma^m,*}^\star u_t + \overline{h}_{\sigma^m,*}^{f,\star} + (s_{t,*}^\star + c_t^{\sigma^m})\mathbb{1}, \quad (13a)$$

**Algorithm 2:** Size Reduction of the Inferred Formula
**Input:** Inferred formula set $E_{Inf} \in \{E_{LTL}, E_{\mathcal{PI}}\}$
1 $E'_{Inf} = \emptyset$
2 **if** $E_{Inf} \neq \emptyset$ **then**
3    **for** $i = 1, \ldots, |E_{Inf}|$ **do**
4       **for** $i' = 1, \ldots, |E_{Inf}|$ **do**
5          **if** $i \neq i'$ and $\hat{\phi}^i \to \phi^{i'}$ **then**
6             **if** $\phi^{i'} \to \phi^i$ and $\phi^{i'} \in E'_{Inf}$ **then**
7                Continue
8             **else**
9                Add $\phi^i$ to $E'_{Inf}$
10                Break
11             **end if**
12          **end if**
13       **end for**
14    **end for**
15 $E_{Red} = E_{Inf}/E'_{Inf}$
16 **if** $E_{Red} \neq \emptyset$ **then**
17    $\psi_I \leftarrow \bigvee_{i=1,\ldots,|E_{Red}|} \phi^i$
18 **else**
19    $\psi_I \leftarrow \phi_I$
20 **end if**
21 **return** $\psi_I$

$$x^\star_{t+1} \geq \underline{A}^\star_{\sigma^m,*} x^\star_t + \underline{B}^\star_{\sigma^m,*} u_t + \overline{h}^{f,\star}_{\sigma^m,*} - (s^\star_{t,*} + c^{\sigma^m}_t)\mathbb{1}, \quad (13b)$$

$$S^{x,\star}_{\sigma^m,*} x^\star_t + S^{u,\star}_{\sigma^m,*} u_t \leq \beta^\star_{\sigma^m,*} + (s^\star_{t,*} + c^{\sigma^m}_t)\mathbb{1}, \quad (13c)$$

$$y_t = Cx^\star_t + v^\star_t, \quad (13d)$$

$$x^{c,\star}_{t,\ell} \leq \overline{H}^\star_{\sigma^{s,\ell},\dagger} z^\star_{t,\ell} + \overline{h}^{g,\star}_{\sigma^{s,\ell},\dagger} + (\tilde{s}^\star_{t,\dagger} + c^{\sigma^{s,\ell}}_t)\mathbb{1}, \quad (13e)$$

$$x^{c,\star}_{t,\ell} \geq \underline{H}^\star_{\sigma^{s,\ell},\dagger} z^\star_{t,\ell} + \underline{h}^{g,\star}_{\sigma^{s,\ell},\dagger} - (\tilde{s}^\star_{t,\dagger} + c^{\sigma^{s,\ell}}_t)\mathbb{1}, \quad (13f)$$

$$S^{z,\star}_{\sigma^{s,\ell},\dagger} z^\star_{t,\ell} \leq \beta^\star_{\sigma^{s,\ell},\dagger} + (\tilde{s}^\star_{t,\dagger} + c^{\sigma^{s,\ell}}_t)\mathbb{1}, \quad (13g)$$

$$a^\star_{t,*} \in \{0,1\}, \tilde{a}^\star_{t,\dagger} \in \{0,1\}, b^{\sigma^m}_t \in \{0,1\}, b^{\sigma^{s,j}}_t \in \{0,1\}, \quad (13h)$$

$$\begin{aligned}&\text{SOS-1:}(a^\star_{t,*}, s^\star_{t,*}), \text{SOS-1:}(\tilde{a}^\star_{t,\dagger}, \tilde{s}^\star_{t,\dagger}),\\&\text{SOS-1:}(b^{\sigma^m}_t, c^{\sigma^m}_t), \text{SOS-1:}(b^{\sigma^{s,j}}_t, c^{\sigma^{s,j}}_t),\end{aligned} \quad (13i)$$

$$\sum_{\xi=1}^{q^{f,\star}_{\sigma^m_t}} a^\star_{t,\xi} = 1, \sum_{\xi=1}^{q^{g,\star}_{\sigma^{s,j}_\ell}} \tilde{a}^\star_{t,\xi} = 1, \sum_{\sigma^m = e_1}^{e_{|\Sigma_m|}} b^{\sigma^m}_t = 1, \quad (13j)$$

$$w^\star_t \in \mathcal{W}, v^\star_t \in \mathcal{V}, u_t \in \mathcal{U}, x^\star_t \in \mathcal{X}, \quad (13k)$$

$$\{\sigma^m_t, \sigma^{s,j}_t\}^{\overline{t}}_{t=\underline{t}} \in V_T(\phi^\star_I), \quad (13l)$$

where $s^\star_{t,*}$, $\tilde{s}^\star_{t,\dagger}$ are slack variables, $c^{\sigma^m}_t$ is vector of slack variables, $c^{\sigma^m}_t = 1$ corresponds to $\sigma^m_t$ being true and 0 otherwise, $\underline{t} = t_0 - T_m + 1$ and $\overline{t} = t_0 + T + T_m - 1$ determines the horizon for $\sigma^m_t$, $\sigma^{s,j}_t$, where $T_m$ is the maximum length of sampled traces (data)[2], the set of valid subtraces $V_T(\phi^\star_I)$ is constructed recursively using encoding of the LTL formula $\phi^\star_I$ according to Section II (cf. using (2)–(8)) and SOS-1 refers to Special Ordered Set of type 1 (i.e., at most one member of the set can be non-zero [69]).

*Proof.* $a^\star_{t,*} = 1$ and $e^{\sigma^m}_t = 1$ imply that (13a)–(13c) hold, since the SOS-1 constraints in (13i) ensure that $s^\star_{t,*} = 0$ and $c^{\sigma^m}_t = 0$ correspondingly. On the contrary, if $a^\star_{t,*} = 0$ and/or $b^{\sigma^m}_t = 0$, it means that $s^\star_{t,*}$ and/or $c^{\sigma^m}_t$ are free and then (13a)–(13c) hold trivially. Similarly, $\tilde{a}^\star_{t,\dagger} = 1$ and $b^{\sigma^{s,j}}_t = 1$, or $\tilde{a}^\star_{t,\dagger} = 0$ and/or $b^{\sigma^{s,j}}_t = 0$ imply that (13e)–(13g) hold. In addition, (13j) ensures that, at each time step $t$, only one partition is valid for each of the state and output equations, and only one switching signal/mode (for $\sigma^m \in \Sigma_m$) is valid. Thus, if the above problem is infeasible, it means that there exists no common behavior that is satisfied by both models, i.e., $\hat{\mathcal{S}}^i \cap \hat{\mathcal{S}}^j = \emptyset$; hence, the pair of models is distinguishable from each other. □

The detection time $T_0$ for all models can then be selected as the maximum value of $T$ for every pair of the given model.

Next, we introduce a data-driven model invalidation algorithm (cf. Algorithm 3) that allows us to rule out/eliminate all models that are incompatible with newly observed data, i.e., to address Problem 3. If not all model pairs are $T$-distinguishable (according to Theorem 2), Algorithm 3 will still return a set of all models that are consistent with the input-mode-output data up to the current time step $t_c$. Specifically, we will solve Problem 3 to check if the newly observed data are consistent with the learned/inferred model-task pairs: In Algorithm 3, we initially verify in Line 4 if the newly observed mode sequence/trace satisfies the inferred formula $\phi^l_I$, and if it does, we proceed to check the compatibility of the learned models $\mathcal{H}^l$ with newly observed trajectory data using the following result in Line 9:

**Theorem 3.** *Given a learned constrained piecewise affine inclusion model $\mathcal{H}^l$ and a (newly observed) length-$T$ input-mode-output sequence $\{u_t, \sigma_t, y_t\}^{t_c}_{t=t_c-T+1}$ at time $t_c$, the model is invalidated if the following problem is infeasible for any $t_c \in \mathbb{Z}^\infty_0$:*

Find $x_t, v_t, s_{t,*}, a_{t,*}, s_{t,\dagger}, a_{t,\dagger}$

$s.t. \forall t \in \mathbb{Z}^{t_c+T-1}_{t_c}, * \in \mathbb{Z}^{q^{f,l}_{\sigma^m_t}}_1, k \in \{\ell \mid \sigma^{s,\ell}_t = 1\}, \dagger \in \mathbb{Z}^{q^{g,l}_{\sigma^{s,k}_t}}_1:$

$$x_{t+1} \leq \overline{A}^l_{\sigma^m_t,*} x_t + \overline{B}^l_{\sigma^m_t,*} u_t + \overline{h}^{f,l}_{\sigma^m_t,*} + s_{t,*}\mathbb{1},$$

$$x_{t+1} \geq \underline{A}^l_{\sigma^m_t,*} x_t + \underline{B}^l_{\sigma^m_t,*} u_t + \underline{h}^{f,l}_{\sigma^m_t,*} - s_{t,*}\mathbb{1},$$

$$S^{x,l}_{\sigma^m_t,*} x_t + S^{u,l}_{\sigma^m_t,*} u_t \leq \beta^l_{\sigma^m_t,*} + s_{t,*}\mathbb{1},$$

$$y_t = Cx_t + v_t, \; v_t \in \mathcal{V}, \; x_t \in \mathcal{X},$$

$$x^c_{t,k} \leq \overline{H}^l_{\sigma^{s,k}_t,\dagger} z_{t,k} + \overline{h}^{g,l}_{\sigma^{s,k}_t,\dagger} + \tilde{s}_{t,\dagger}\mathbb{1},$$

$$x^c_{t,k} \geq \underline{H}^l_{\sigma^{s,k}_t,\dagger} z_{t,k} + \underline{h}^{g,l}_{\sigma^{s,k}_t,\dagger} - \tilde{s}_{t,\dagger}\mathbb{1},$$

$$S^{z,l}_{\sigma^{s,k}_t,\dagger} z_{t,k} \leq \beta^l_{\sigma^{s,k}_t,\dagger} + \tilde{s}_{t,\dagger}\mathbb{1},$$

$$a_{t,*} \in \{0,1\}, \; \text{SOS-1:}(a_{t,*}, s_{t,*}), \; \sum_{\xi=1}^{q^{f,l}_{\sigma^m_t}} a_{t,\xi} = 1,$$

$$\tilde{a}_{t,\dagger} \in \{0,1\}, \; \text{SOS-1:}(\tilde{a}_{t,\dagger}, \tilde{s}_{t,\dagger}), \; \sum_{\xi=1}^{q^{g,l}_{\sigma^{s,k}_t}} \tilde{a}_{t,\xi} = 1.$$

*Proof.* The construction follows similar steps to Theorem 2, but with only one model and a given/measured mode sequence. □

If either of the checks is infeasible, then that learned model-task pair can be ruled out as incompatible.

However, when the size of the inferred formulas is large (even after applying the reduction approach using Algorithm 2 in Section V-B.3), the resulting optimization problems in

---
[2] Although LTL formulas are usually unbounded, we can use the length of the data used to infer the LTL formula (in Section V) to determine $\underline{t}$ and $\overline{t}$ that bidirectionally extend the time horizon based on the "LTL bound."

**Algorithm 3:** Model Discrimination with Length $T$

**Input:** Models $\{\mathcal{G}^l\}_{l=1}^{N_m}$, Input-Output Sequence = $\{u_t, \sigma_t, y_t\}_{t=t_0-T+1}^{t_0}$

1 **function** findModel($\{\mathcal{G}^l\}_{l=1}^{N_m}, \{u_t, \sigma_t, y_t\}_{t=t_0-T+1}^{t_0}$)
2    valid $\leftarrow \{\mathcal{G}^l\}_{l=1}^{N_m}$;
3    **for** $l = 1 : N_m$ **do**
4       Check $\sigma \models \phi_I^l$;
5       **if** *infeasible* **then**
6          Remove $l$ from valid;
7          Continue;
8       **end if**
9       Check Feasibility of Theorem 3;
10      **if** *infeasible* **then**
11         Remove $l$ from valid;
12      **end if**
13   **end for**
14   **return** valid

both Theorems 2 and 3 may also be large and this results in high computational cost and large memory requirements. To remedy this, in practice, since the $\bigvee$ condition can be considered separately, we can instead consider all possible combinations of the subformulas in parallel and the maximum $T$ among all these combinations will be considered as the solution to Problem 2, while for Problem 3, the problem in Theorem 3 is infeasible if the subproblems with all combinations of subformulas are all infeasible. For instance, for a model-task pair with inferred formulas $\phi_I^i = \bigvee_{k=1}^{k=|E_i|} \phi_k^i$ and $\phi_I^j = \bigvee_{k=1}^{k=|E_j|} \phi_k^j$, we can consider all combinations of subformulas of these pairs of formulas, i.e., the subformula pairs $(\phi_k^i, \phi_{k'}^j), \forall k = \mathbb{Z}_1^{|E_i|}, k' \in \mathbb{Z}_1^{|E_j|}$.

## VII. SIMULATIONS

The simulations in Subsections VII-VII-A and VII-VII-B are implemented in MATLAB 2021a with the Gurobi optimization solver [69] on a 2.2 GHz machine with 16 GB RAM.

### A. Single Link Robot Arm Link

First, we tested our algorithms using a single link robot arm example with the following dynamics [70]:

$$\ddot{x}(t) = -\frac{m^k g l^k}{J^k}\sin(x(t)) - \frac{D^k}{J^k}\dot{x}(t) + \frac{1}{J^k}u(t), \quad (15)$$

where $x(t)$ is the angle of the arm, $u(t)$ is the control input, and for each discrete state/mode $q^k = (m^k, l^k, J^k, D^k)$, $m^k$ is the mass, $l^k$ is the length of the arm that can extend and retract (making the robot arm a switched system), $J^k$ is the moment of inertia, and $D^k$ is the damping coefficient. These parameters $q^k$ change depending on the angle parameter $x(t)$. We discretized the system with sampling time $\delta t = 0.1s$ and select the $u(t) = 0.5\cos(t)$.

For the first model, the ground truth formula for the task specification is $\phi^1 = (\sigma^{m,1} \rightarrow (\neg\sigma^{m,2} \mathbf{U} \sigma^{m,1})) \vee \sigma^{m,3}$ where $\sigma^{m,k}$ is an atomic proposition for $\sigma_t^m = e_k$ and $q^k$ can be one of three values: $q^1 = (1, 1, 2, 2)$, $q^2 = (1, 1.5, 3, 475, 2)$, or $q^3 = (1, 2, 8, 2)$. The second model is with the ground truth formula $\phi^2 = (\mathbf{G}\,\sigma^{m,1} \vee \sigma^{m,2}) \vee$

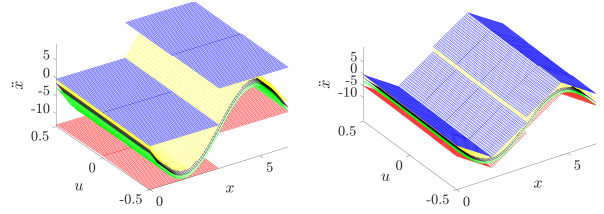

(a) Model with 4 partitions    (b) Model with 16 partitions

Fig. 4: Illustration of the learned over-approximation model with different number of partitions for the function in (15) with $q^k = (1, 1, 2, 2)$ in Example A.
$\mathbf{G}\,\neg\sigma^{m,3}$ and $q^k$ can be one of three values: $q^1 = (2, 1, 2, 2)$, $q^2 = (2, 1.5, 3, 475, 2)$, or $q^3 = (2, 2, 8, 2)$.

We first generated 200 random data points for each mode $\begin{bmatrix} \dot{x}_{t+1} \\ x_{t+1} \end{bmatrix} = f^l\left(\begin{bmatrix} \dot{x}_t \\ x_t \end{bmatrix}, u_t\right)$ of both time-discretized models and used this data to learn the inclusion/over-approximated dynamics model as shown in Fig. 4, where the true model is included/contained within the hyperplanes of learned dynamics models, as expected, and the more partitions we consider, the better the model accuracy will be.

For each model, we generated 50 traces with three modes, all with a maximum length of 20 time steps. To infer $\phi_I^1$, we used the prior information $\mathcal{PI}$ in Figure 5 (left) and obtained $\phi_I^1 = (\sigma^{m,1}) \vee ((\sigma^{m,1} \mathbf{U} \sigma^{m,1}) \vee \sigma^{m,3}) \vee (\sigma^{m,1} \rightarrow (\sigma^{m,1} \mathbf{U} \sigma^{m,1})) \vee (((\neg\sigma^{m,2} \mathbf{U} \sigma^{m,1}) \wedge \sigma^{m,1}) \vee \sigma^{m,3}) \vee ((\sigma^{m,1} \rightarrow (\neg\sigma^{m,2} \mathbf{U} \sigma^{m,1})) \vee \sigma^{m,3})$. To infer $\phi_I^2$, we used the prior information $\mathcal{PI}$ depicted in Figure 5 (right) and obtained $\phi_I^2 = (\sigma^{m,1} \vee \sigma^{m,2}) \vee (((\mathbf{G}\,\sigma^{m,1}) \vee \sigma^{m,2}) \vee (\mathbf{G}(\neg\sigma^{m,3}))) \vee (((\sigma^{m,1} \vee \sigma^{m,2}) \vee \mathbf{G}(\neg\sigma^{m,3})) \vee (\neg\sigma^{m,3})) \vee ((\sigma^{m,1} \vee \sigma^{m,2}) \vee \mathbf{G}(\neg\sigma^{m,3}))$. We reduced the size of the inferred specifications using Algorithm 2, and obtained $\psi_I^1 = \sigma^{m,1} \rightarrow (\sigma^{m,1} \mathbf{U} \sigma^{m,1})$ and $\psi_I^2 = ((\sigma^{m,1} \vee \sigma^{m,2}) \vee \mathbf{G}(\neg\sigma^{m,3})) \vee (\neg\sigma^{m,3})$.

From Table I, we can see that the guaranteed detection time are the same for all cases, and 'with formulas' case requires more computational time when compared to the 'without formulas' case due to the extra integer constraints in the optimization problem when considering the inferred LTL formulas. However, since finding the detection time $T$ is an off-line process, the CPU time matters less than the CPU time for the model discrimination, which is an online process. As shown in Table II, the mean CPU time was lower when using both learned/inferred system dynamics and LTL formulas than only using the learned dynamics, while using only the inferred LTL formulas results in only a discrimination rate of 20%.

### B. Dubins Car

In this example, we consider a lane change scenario with Dubins Car model [71] as the dynamics model of the vehicle:

$$p_{x,t+1} = p_{x,t} + u_s\cos(\theta_t)\delta t + w_{px,t},$$
$$p_{y,t+1} = p_{y,t} + u_s\sin(\theta_i)\delta t + w_{py,t},$$
$$\theta_{t+1} = \theta_t + \frac{u_s}{L}\tan(u_\phi)\delta t + w_{\theta,t},$$

where the system states are $p_x$ and $p_y$ that represent the $(x, y)$-position of the agent and $\theta$ as the heading angle of the agent. $L$ is the length between the front and rear tires and is set to $1.5m$, $u_s$ is the speed of the agent and is assumed to be $1\,m/s$, and sampling time $\delta t$ is set to $0.1s$. The vehicle utilizes for feedback control according to the proportional control law: $u_\phi = K^k(\theta_{desired} - \theta)$.

TABLE I: Comparison of the guaranteed detection times $T$ and CPU times ($s$) for model discrimination (cf. Theorem 2) with and without inferred LTL formulas in Example A.

|  | $T$ | CPU Time ($s$) |
|---|---|---|
| (i) Without formulas | 17 | 165.02 |
| (ii) With formulas | 17 | 403.20 |

TABLE II: Comparison of model discrimination results (with Algorithm 3) for 20 different sequences with length $T = 17$ in Example A.

|  | Mean CPU Time ($s$) | Discrimination Rate |
|---|---|---|
| (i) Dynamics with formulas | 2.95 | 100% |
| (ii) Only dynamics | 3.13 | 100% |
| (iii) Only formulas (reduced) | 0.02 | 20% |
| (iv) Only formulas (all) | 0.42 | 20% |

TABLE III: Comparison of the guaranteed detection times $T$ and CPU times ($s$) for model discrimination (cf. Theorem 2) with and without inferred LTL formulas in Example B.

|  | $T$ | CPU Time ($s$) |
|---|---|---|
| (i) Without formula | 15 | 132.73 |
| (ii) With formula | 15 | 373.10 |

TABLE IV: Comparison of model discrimination results (with Algorithm 3) for 20 different sequences with length $T = 15$ in Example B.

|  | Mean CPU Time ($s$) | Discrimination Rate |
|---|---|---|
| (i) Dynamics with formulas | 2.63 | 100% |
| (ii) Only dynamics | 3.02 | 100% |
| (iii) Only formulas (reduced) | 0.07 | 50% |
| (iv) Only formulas (all) | 1.02 | 50% |

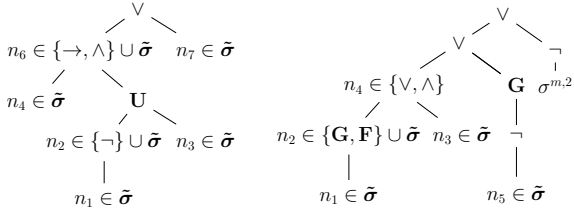

Fig. 5: The DAG structure representing the given $\mathcal{PI}$ for the model-task pair $(\mathcal{G}^1, \phi^1)$ (left) and $(\mathcal{G}^2, \phi^2)$ (right) in Example A with $\tilde{\boldsymbol{\sigma}} \triangleq \{\sigma^{m,1}, \sigma^{m,2}, \sigma^{m,3}\}$.

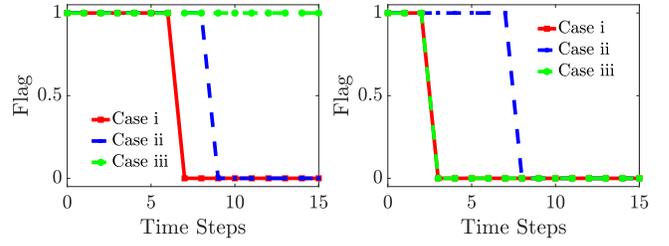

Fig. 6: Model discrimination results (with Algorithm 3) using two different newly observed sequences to invalidate a non-ground truth/false model in Example B under three different cases, $j \in \{\text{i, ii, iii}\}$ where Case i leverages both the learned/inferred system dynamics and formulas, Case ii only leverages the learned system dynamics and Case iii only leverages the inferred formulas. Flag $j$ is 1 when the false model is not (yet) invalidated and is 0 if invalidated.

For the first model, the ground truth formula for the system is $\phi^1 = (\sigma^{m,1}\,\mathbf{U}\,\sigma^{m,2}) \wedge \sigma^{s,1}$ and $K^k$ can take two values: $K^1 = 0.5$ and $K^2 = 0.1$ for $\sigma^{m,1}$ and $\sigma^{m,2}$, respectively, with $\theta_{desired} = 60°$, while $\sigma^{m,k}$ is an atomic proposition for $\sigma_t^m = e_k$ and $\sigma^{s,1}$ is an atomic proposition for the state constraint $p_y = \frac{e^{(p_x+6)}}{1+e^{(p_x+6)}}$. For the second model, the ground truth formula for the system is $\phi^2 = (\sigma^{m,1}\,\mathbf{U}\,\sigma^{m,2}) \vee \sigma^{s,1}$ and $K^k$ can take two values: $K^1 = -0.5$ and $K^2 = -0.1$ for $\sigma^{m,1}$ and $\sigma^{m,2}$, respectively, with $\theta_{desired} = -60°$, while $\sigma^{m,k}$ is an atomic proposition for $\sigma_t^m = e_k$ and $\sigma^{s,1}$ is an atomic proposition for the state constraint $p_y = -\frac{e^{(p_x+6)}}{1+e^{(p_x+6)}}$.

We first generated 200 random data points for each mode of both models to learn the inclusion/over-approximation model of the system dynamics (not depicted for brevity). Further, we generated 50 traces with three modes, all with a maximum length of 20 time steps. Using the prior information $\mathcal{PI}$ in Figure 7 (left), we obtained $\phi_I^1 = (\sigma^{s,1}) \vee (((\mathbf{G}\,\sigma^{s,1})\,\mathbf{U}(\mathbf{G}\,\sigma^{s,1})) \vee \sigma^{s,1}) \vee ((\sigma^{m,1}\,\mathbf{U}\,\sigma^{m,2}) \wedge \sigma^{s,1}) \vee ((\sigma^{m,2}\,\mathbf{U}(\mathbf{G}\,\sigma^{s,1})) \vee \sigma^{s,1})$. For the second model, using the prior information in Figure 7 (right), we obtained $\phi_I^2 = (\sigma^{m,1} \vee (\sigma^{m,2}\,\mathbf{U}\,\sigma^{m,2})) \vee ((\sigma^{m,1}\,\mathbf{U}\,\sigma^{m,2}) \vee \sigma^{s,1}) \vee (((\mathbf{G}\,\sigma^{s,1})\,\mathbf{U}\,\sigma^{m,2}) \vee \sigma^{m,1}) \vee (\sigma^{m,1}\,\mathbf{U}\,\sigma^{m,2})$. We reduced the size of the inferred formulas using Algorithm 2, and obtained $\psi_I^1 = (\sigma^{m,2}\,\mathbf{U}(\mathbf{G}\,\sigma^{s,1})) \vee \sigma^{s,1}$ and $\psi_I^2 = ((\sigma^{m,1}\,\mathbf{U}\,\sigma^{m,2}) \vee \sigma^{s,1}) \vee (((\mathbf{G}\,\sigma^{s,1})\,\mathbf{U}\,\sigma^{m,2}) \vee \sigma^{m,1})$.

Similar to the first example, we can see that the guaranteed detection time are the same for all cases from Table III, while in Table IV, the mean CPU time can be observed to be lower when using both the learned/inferred system dynamics and LTL formulas than when only using the learned dynamics and the discrimination rate is only 50% if only the inferred formulas are used. We also depicted model discrimination results (with Algorithm 3) in Figure 6, where two sampled sequences from the ground truth model are used to invalidate learned model-task pairs from a different non-ground truth model. In the figure, we can observe that when using both the inferred system dynamics and LTL formulas (Case i), the non-ground truth model is always invalidated earlier (Flag is 0) than either using only the learned dynamics (Case ii) or only the inferred formulas (Case iii). Moreover, in the left instance, when using only the inferred formulas, the non-ground truth model was not able to be invalidated.

## VIII. CONCLUSION

To address the challenge of data-driven model discrimination for unknown switched systems with unknown LTL specifications, this paper first proposed a data-driven set-membership method that over-approximates the unknown dynamics. Then, the paper introduced a specification inference method that guarantees that the ground truth LTL is among/included in the inferred formulas, along with a method for reducing the size of the inferred formula when prior information is available. Next, we introduced an optimization-based algorithm to analyze model detectability from noisy, finite data, and a model discrimination algorithm that can rule out learned/inferred model-task pairs that are inconsistent with new observations at run time. The effectiveness of the proposed methods was demonstrated through several illustrative examples. The results indicated that the joint learning/inferring and use of both system dynamics and temporal logic specifications can significantly accelerate the data-driven model discrimination process. Future work will extend the proposed approach to consider inequality constraints in the system model and missing/dropped data.

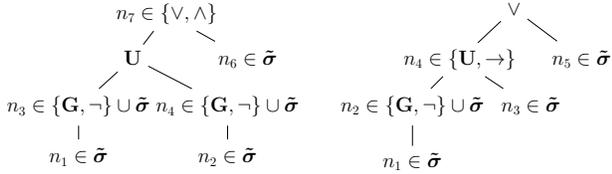

Fig. 7: The DAG structure representing the given $\mathcal{PI}$ for the model-task pair $(\mathcal{G}^1, \phi^1)$ (left) and $(\mathcal{G}^2, \phi^2)$ (right) in Example B with $\tilde{\boldsymbol{\sigma}} \triangleq \{\sigma^{m,1}, \sigma^{m,2}, \sigma^{s,1}\}$.